\documentclass[10pt,twocolumn,letterpaper]{article}
\pdfoutput=1
\usepackage{subcaption}
\usepackage{cvpr}
\usepackage{times}
\usepackage{epsfig}
\usepackage{graphicx}
\usepackage{amsmath}
\usepackage{amssymb}
\usepackage{booktabs}
\newcommand{\firstrank}[1]{\textcolor{red}{\textbf{#1}}}
\newcommand{\secondrank}[1]{\textcolor{blue}{\textit{#1}}}
\newenvironment{packed_enum}{
\begin{enumerate}
  \setlength{\itemsep}{2.5pt}
  \setlength{\parskip}{0pt}
  \setlength{\parsep}{0pt}
}{\end{enumerate}}

\usepackage[pagebackref=true,breaklinks=true,letterpaper=true,colorlinks,bookmarks=false]{hyperref}

\cvprfinalcopy 

\makeatletter\renewcommand\paragraph{\@startsection{paragraph}{4}{\z@}
  {.5em \@plus1ex \@minus.2ex}{-.2em}{\normalfont\normalsize\bfseries}}\makeatother
\def\Strain{S_\mathrm{train}}
\def\Smeta{S_\mathrm{meta}}
\def\Stest{S_\mathrm{test}}
\def\BiLSTM{\mathrm{BiLSTM}}
\def\AttLSTM{\mathrm{AttLSTM}}
\def\ngen{n_\mathrm{gen}}
\def\naug{n_\mathrm{aug}}
\def\Cbase{C_\mathrm{base}}
\def\Cnovel{C_\mathrm{novel}}

\def\cvprPaperID{****} 

\ifcvprfinal\fi
\begin{document}

\title{Low-Shot Learning from Imaginary Data}

\author{
Yu-Xiong Wang$^{1,2}$ \quad Ross Girshick$^1$ \quad Martial Hebert$^2$ \quad Bharath Hariharan$^{1,3}$\\[2mm]
$^1$Facebook AI Research (FAIR) \qquad $^2$Carnegie Mellon University \qquad $^3$Cornell University
}

\maketitle

\begin{abstract}
Humans can quickly learn new visual concepts, perhaps because they can easily visualize or imagine what novel objects look like from different views.
Incorporating this ability to hallucinate novel instances of new concepts might help machine vision systems perform better low-shot learning, i.e., learning concepts from few examples.
We present a novel approach to low-shot learning that uses this idea.
Our approach builds on recent progress in meta-learning (``learning to learn'') by combining a meta-learner with a ``hallucinator'' that produces additional training examples, and optimizing both models jointly.
Our hallucinator can be incorporated into a variety of meta-learners and provides significant gains: up to a 6 point boost in classification accuracy when only a single training example is available, yielding state-of-the-art performance on the challenging ImageNet low-shot classification benchmark.
\end{abstract}

\section{Introduction}
The accuracy of visual recognition systems has grown dramatically.
But modern recognition systems still need thousands of examples of each class to saturate performance.
This is impractical in cases where one does not have enough resources to collect large training sets or that involve rare visual concepts.
It is also unlike the human visual system, which can learn a novel visual concept from even a single example~\cite{Schmidt2009}.
This challenge of learning new concepts from very few labeled examples, often called \emph{low-shot} or \emph{few-shot learning}, is the focus of this work.

Many recently proposed approaches to this problem fall under the umbrella of \emph{meta-learning}~\cite{Thrun1998}.
Meta-learning methods train a \emph{learner}, which is a parametrized function that maps labeled training sets to classifiers.
Meta-learners are trained by \emph{sampling} small training sets and test sets from a large universe of labeled examples, feeding the sampled training set to the learner to get a classifier, and then computing the loss of the classifier on the sampled test set.
These methods directly frame low-shot learning as an optimization problem.

\begin{figure}[!t]
\centering
\includegraphics[width=0.78\linewidth]{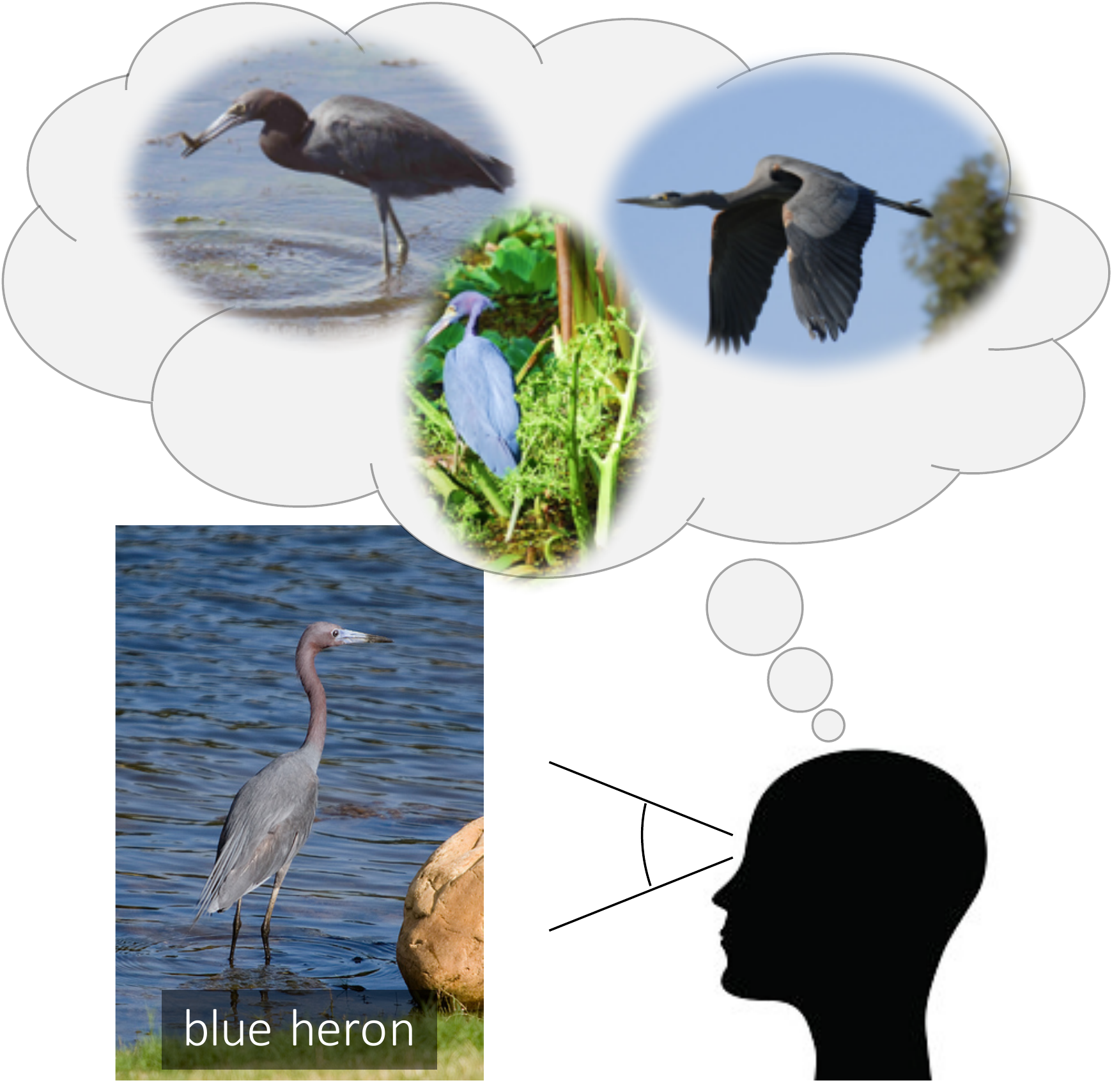}
\caption{Given a single image of a novel visual concept, such as a blue heron, a person can visualize what the heron would look like in other poses and different surroundings. If computer recognition systems could do such hallucination, they might be able to learn novel visual concepts from less data.}
\vspace{-0.5cm}
\label{fig:1}
\end{figure}

However, generic meta-learning methods treat images as black boxes, ignoring the structure of the visual world.
In particular, many modes of variation (for example camera pose, translation, lighting changes, and even articulation) are shared across categories.
As humans, our knowledge of these shared modes of variation may allow us to visualize what a novel object might look like in other poses or surroundings (Figure~\ref{fig:1}).
If machine vision systems could do such ``hallucination'' or ``imagination'', then the hallucinated examples could be used as additional training data to build better classifiers.

Unfortunately, building models that can perform such hallucination is hard, except for simple domains like handwritten characters~\cite{MillerCVPR2000}.
For general images, while considerable progress has been made recently in producing realistic samples, most current generative modeling approaches suffer from the problem of mode collapse~\cite{SalimansNIPS2016}: they are only able to capture some modes of the data.
This may be insufficient for low-shot learning since one needs to capture many modes of variation to be able to build good classifiers. Furthermore, the modes that are useful for classification may be different from those that are found by training an image generator.
Prior work has tried to avoid this limitation by explicitly using pose annotations to generate samples in novel poses~\cite{DixitCVPR2017}, or by using carefully designed, but brittle, heuristics to ensure diversity~\cite{HariharanICCV2017}.

Our key insight is that the criterion that we should aim for when hallucinating additional examples is neither diversity nor realism. Instead, the aim should be to hallucinate examples that are \emph{useful} for learning classifiers. Therefore, we propose a new method for low-shot learning that directly learns to hallucinate examples that are useful for classification by the end-to-end optimization of a classification objective that includes data hallucination in the model.

We achieve this goal by unifying meta-learning with hallucination.
Our approach trains not just the meta-learner, but also a \emph{hallucinator}: a model that maps real examples to hallucinated examples. 
The few-shot training set is first fed to the hallucinator; it produces an expanded training set, which is then used by the learner.
Compared to plain meta-learning, our approach uses the rich structure of shared modes of variation in the visual world. We show empirically that such hallucination adds a significant performance boost to two different meta-learning methods~\cite{VinyalsNIPS2016,SnellArxiv2017}, providing up to a 6 point improvement when only a single training example is available.
Our method is also agnostic to the choice of the meta-learning method, and provides significant gains irrespective of this choice. It is precisely the ability to leverage standard meta-learning approaches \emph{without any modifications} that makes our model simple, general, and very easy to reproduce. 
Compared to prior work on hallucinating examples, we use no extra annotation and significantly outperform hallucination based on brittle heuristics~\cite{HariharanICCV2017}. 
We also present a novel meta-learning method and discover and fix flaws in previously proposed benchmarks.

\vspace{-0.2cm}
\section{Related Work}
\vspace{-0.1cm}
Low-shot learning is a classic problem~\cite{ThrunNIPS1996}.
One class of approaches builds generative models that can share priors across categories~\cite{FeiFeiTPAMI2006, Salakhutdinov2012,george2017generative}.
Often, these generative models have to be hand-designed for the domain, such as strokes~\cite{LakeNIPS2013, LakeScience2015} or parts~\cite{WongICCV2015} for handwritten characters.
For more unconstrained domains, while there has been significant recent progress~\cite{RezendeICML2014, GoodfellowNIPS2014, RedfordICLR2016}, modern generative models still cannot capture the entirety of the distribution~\cite{SalimansNIPS2016}.

Different classes might not share parts or strokes, but may still share modes of variation, since these often correspond to camera pose, articulation, \etc.
If one has a probability density on transformations, then one can generate additional examples for a novel class by applying sampled transformations to the provided examples~\cite{MillerCVPR2000, DixitCVPR2017, HariharanICCV2017}.
Learning such a density is easier for handwritten characters that only undergo 2D transformations~\cite{MillerCVPR2000}, but much harder for generic image categories.
Dixit \etal~\cite{DixitCVPR2017} tackle this problem by leveraging an additional dataset of images labeled with pose and attributes; this allows them to learn how images transform when the pose or the attributes are altered.
To avoid annotation, Hariharan and Girshick~\cite{HariharanICCV2017}  try to transfer transformations from a pair of examples from a known category to a ``seed" example of a novel class.
However, learning to do this transfer requires a carefully designed pipeline with many heuristic steps.
Our approach follows this line of work, but learns to do such transformations in an end-to-end manner, avoiding both brittle heuristics and expensive annotations.

Another class of approaches to low-shot learning has focused on building feature representations that are invariant to intra-class variation.
Some work tries to share features between seen and novel classes~\cite{BartCVPR2005,wang2016learningfrom} or incrementally learn them as new classes are encountered~\cite{OpeltCVPR2006}.
Contrastive loss functions~\cite{HadsellCVPR2006, KochICMLW2015} and variants of the triplet loss~\cite{TaigmanCVPR2015, SchroffCVPR2015, FinkNIPS2005} have been used for learning feature representations suitable for low-shot learning; the idea is to push examples from the same class closer together, and farther from other classes.
Hariharan and Girshick~\cite{HariharanICCV2017} show that one can encourage classifiers trained on small datasets to match those trained on large datasets by a carefully designed loss function.
These representation improvements are orthogonal to our approach, which works with any features.

More generally, a recent class of methods tries to frame low-shot learning itself as a ``learning to learn'' task, called meta-learning~\cite{Thrun1998}.
The idea is to directly train a parametrized mapping from training sets to classifiers.
Often, the learner embeds examples into a feature space.
It might then accumulate statistics over the training set using recurrent neural networks (RNNs)~\cite{VinyalsNIPS2016, RaviICLR2017}, memory-augmented networks~\cite{SantoroICML2016}, or multilayer perceptrons (MLPs)~\cite{EdwardsICLR2017}, perform gradient descent steps to finetune the representation~\cite{FinnICML2017}, and/or collapse each class into prototypes~\cite{SnellArxiv2017}.
An alternative is to directly predict the classifier weights that would be learned from a large dataset using few novel class examples~\cite{BertinettoNIPS2016} or from a small dataset classifier~\cite{WangECCV2016,wang2017learning}.
We present a unified view of meta-learning and show that our hallucination strategy can be adopted in any of these methods.

\section{Meta-Learning}
\vspace{-0.1cm}
Let $\mathcal{X}$ be the space of inputs (\eg, images) and $\mathcal{Y}$ be a discrete label space.
Let $\mathcal{D}$ be a distribution over $\mathcal{X} \times \mathcal{Y}$.
Supervised machine learning typically aims to capture the conditional distribution $p(y | x)$ by applying a \emph{learning algorithm} to a parameterized model and a training set $\Strain  = \{(x_i,y_i) \sim \mathcal{D}\}_{i=1}^N$.
At inference time, the model is evaluated on test inputs $x$ to estimate $p(y|x)$.
The composition of the inference and learning algorithms can be written as a function $h$ (a \emph{classification algorithm}) that takes as input the training set and a test input $x$, and outputs an estimated probability distribution $\hat{\mathbf{p}}$ over the labels:
\begin{equation}
\vspace{-0.1cm}
\hat{\mathbf{p}}(x) = h(x, \Strain).
\end{equation}

In low-shot learning, we want functions $h$ that have high classification accuracy even when $\Strain$ is small.
Meta-learning is an umbrella term that covers a number of recently proposed empirical risk minimization approaches to this problem~\cite{WangECCV2016, VinyalsNIPS2016, SnellArxiv2017, FinnICML2017, RaviICLR2017}.
Concretely, they consider \emph{parametrized} classification algorithms $h(\cdot, \cdot; \mathbf{w})$ and attempt to estimate a ``good" parameter vector $\mathbf{w}$, namely one that corresponds to a classification algorithm that can learn well from small datasets.
Thus, estimating this parameter vector can be construed as \emph{meta-learning}~\cite{Thrun1998}.

Meta-learning algorithms have two stages.
The first stage is \emph{meta-training} in which the parameter vector $\mathbf{w}$ of the classification algorithm is estimated.
During meta-training, the meta-learner has access to a large labeled dataset $\Smeta$ that typically contains thousands of images for a large number of classes $C$.
In each iteration of meta-training, the meta-learner samples a \emph{classification problem} out of $\Smeta$.
That is, the meta-learner first samples a subset of $m$ classes from $C$.
Then it samples a small ``training'' set $\Strain$ and a small ``test'' set $\Stest$.
It then uses its current weight vector $\mathbf{w}$ to compute conditional probabilities $h(x, \Strain; \mathbf{w})$ for every point $(x,y)$ in the test set $\Stest$.
Note that in this process $h$ may perform internal computations that amount to ``training'' on $\Strain$.
Based on these predictions, $h$ incurs a loss $L(h(x, \Strain; \mathbf{w}), y)$ for each point in the current $\Stest$.
The meta-learner then back-propagates the gradient of the total loss $\sum_{(x,y) \in \Stest} L(h(x, \Strain; \mathbf{w}), y)$.
The number of classes in each iteration, $m$, and the maximum number of training examples per class, $n$, are hyperparameters.

The second stage is \emph{meta-testing} in which the resulting classification algorithm is used to solve novel classification tasks: for each novel task, the labeled training set and unlabeled test examples are given to the classification algorithm and the algorithm outputs class probabilities.

Different meta-learning approaches differ in the form of $h$.
The data hallucination method introduced in this paper is general and applies to any meta-learning algorithm of the form described above.
Concretely, we will consider the following three meta-learning approaches:
\paragraph{Prototypical networks:}
Snell \etal~\cite{SnellArxiv2017} propose an architecture for $h$ that assigns class probabilities based on distances from class means $\mu_k$ in a learned feature space:
\begin{align}
\vspace{0.1cm}
h(x, \Strain; &\mathbf{w}) = \hat{\mathbf{p}}(x) \\
\hat{p}_k(x) =& \frac{e^{-d(\phi(x; \mathbf{w}_\phi), \mu_k)}}{\sum_j e^{-d(\phi(x; \mathbf{w}_\phi), \mu_j)}}\\
\mu_k =& \frac{\sum_{(x_i,y_i) \in \Strain} \phi(x_i; \mathbf{w}_\phi) \mathbf{I}[y_i=k]} {\sum_{(x_i,y_i) \in \Strain}  \mathbf{I}[y_i=k]}.
\vspace{0.1cm}
\end{align}

\vspace{0.2cm}
Here $\hat{p}_k$ are the components of the probability vector $\hat{\mathbf{p}}$ and $d$ is a distance metric (Euclidean distance in~\cite{SnellArxiv2017}).
The only parameters to be learned here are the parameters of the feature extractor $\mathbf{w}_\phi$.
The estimation of the class means $\mu_k$ can be seen as a simple form of ``learning'' from $\Strain$ that takes place internal to $h$.

\paragraph{Matching networks:}
Vinyals \etal~\cite{VinyalsNIPS2016} argue that when faced with a classification problem and an associated training set, one wants to focus on the features that are useful for \emph{those particular class distinctions}.
Therefore, after embedding all training and test points independently using a feature extractor, they propose to create a \emph{contextual embedding} of the training and test examples using bi-directional long short-term memory networks (LSTMs) and attention LSTMs, respectively.
These contextual embeddings can be seen as emphasizing features that are relevant for the particular classes in question.
The final class probabilities are computed using a soft nearest-neighbor mechanism. More specifically,
\begin{align}
\vspace{0.1cm}
h(x, \Strain; &\mathbf{w}) = \hat{\mathbf{p}}(x) \\
\hat{p}_k(x) =& \frac{\sum_{(x_i,y_i) \in \Strain} e^{-d(f(x), g(x_i))}\mathbf{I}[y_i=k]}{\sum_{(x_i,y_i) \in \Strain} e^{-d(f(x), g(x_i))}}\\
f(x) =& \AttLSTM(\phi(x; \mathbf{w}_\phi), \{g(x_i)\}_{i=1}^N; \mathbf{w}_f)\\
\{g(x_i)\}_{i=1}^N =& \BiLSTM(\{\phi(x_i; \mathbf{w}_\phi)\}_{i=1}^N; \mathbf{w}_g).
\vspace{0.1cm}
\end{align}

\vspace{0.2cm}
Here, again $d$ is a distance metric.
Vinyals \etal used the cosine distance.
There are three sets of parameters to be learned: $\mathbf{w}_\phi, \mathbf{w}_g,$ and $\mathbf{w}_f$.

\paragraph{Prototype matching networks:}
One issue with matching networks is that the attention LSTM might find it harder to ``attend'' to rare classes (they are swamped by examples of common classes), and therefore might introduce heavy bias against them.
Prototypical networks do not have this problem since they collapse every class to a single  class mean.
We want to combine the benefits of the contextual embedding in matching networks with the resilience to class imbalance provided by prototypical networks.

To do so, we collapse every class to its class mean before creating the contextual embeddings of the test examples. Then, the final class probabilities are based on distances to the contextually embedded class means instead of individual examples:

\begin{align}
h(x, \Strain; &\mathbf{w}) = \hat{\mathbf{p}}(x) \\
\hat{p}_k(x) =& \frac{e^{-d(f(x), \nu_k)}}{\sum_j e^{-d(f(x), \nu_j)}}\\
f(x) = & \AttLSTM(\phi(x; \mathbf{w}_\phi), \{\nu_k\}_{k=1}^{|\mathcal{Y}|}; \mathbf{w}_f) \\
\nu_k =& \frac{\sum_{(x_i,y_i) \in \Strain} g(x_i) \mathbf{I}[y_i=k]} {\sum_{(x_i,y_i) \in \Strain}  \mathbf{I}[y_i=k]}\\
\{g(x_i)\}_{i=1}^N =& \BiLSTM(\{\phi(x_i; \mathbf{w}_\phi)\}_{i=1}^N; \mathbf{w}_g).
\end{align}

The parameters to be learned are $\mathbf{w}_\phi, \mathbf{w}_g$, and $\mathbf{w}_f$.
We call this novel modification to matching networks \emph{prototype matching networks}.

\section{Meta-Learning with Learned Hallucination}
\vspace{-0.1cm}
\begin{figure}
\centering
\includegraphics[width=.9\linewidth]{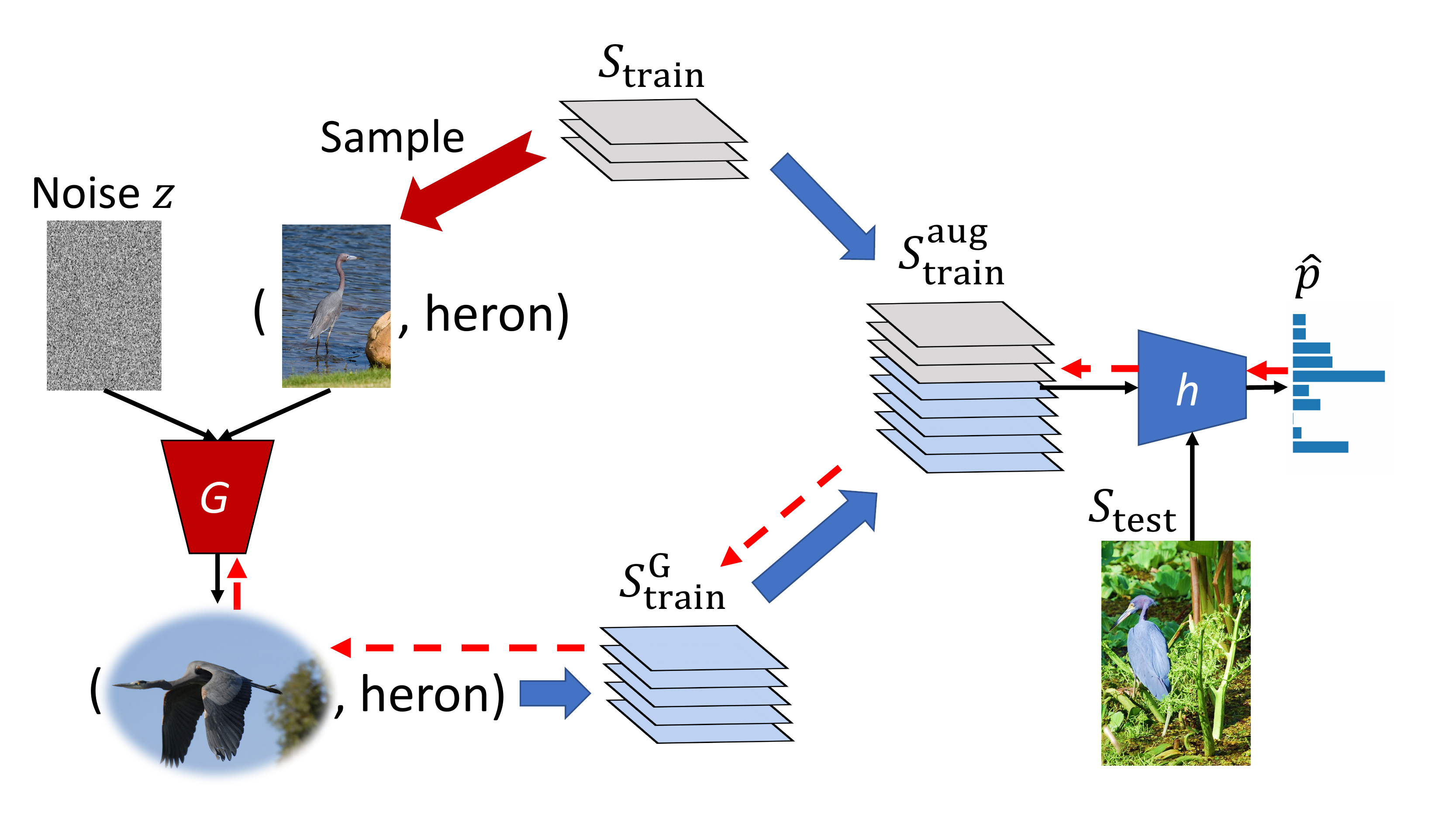}
\vspace{-0.2cm}
\caption{Meta-learning with hallucination. Given an initial training set $\Strain$, we create an augmented training set $\Strain^\mathrm{aug}$ by adding a set of generated examples $\Strain^G$. $\Strain^G$ is obtained by sampling real seed examples and noise vectors $z$ and passing them to a parametric hallucinator $G$. The hallucinator is trained end-to-end along with the classification algorithm $h$. Dotted red arrows indicate the flow of gradients during back-propagation.}
\vspace{-0.2cm}
\label{fig:2}
\end{figure}
We now present our approach to low-shot learning by learning to hallucinate additional examples.
Given an initial training set $\Strain$, we want a way of \emph{sampling} additional hallucinated examples.
Following recent work on generative modeling~\cite{GoodfellowNIPS2014, kingma2014auto}, we will model this stochastic process by way of a \emph{deterministic} function operating on a \emph{noise vector} as input.
Intuitively, we want our hallucinator to take a single example of an object category and produce other examples in different poses or different surroundings.
We therefore write this hallucinator as a function $G(x,z; \mathbf{w}_G)$ that takes a seed example $x$ and a noise vector $z$ as input, and produces a hallucinated example as output.
The parameters of this hallucinator are $\mathbf{w}_G$.

We first describe how this hallucinator is used in meta-testing, and then discuss how we train the hallucinator.
\paragraph{Hallucination during meta-testing:}
During meta-testing, we are given an initial training set $\Strain$.
We then hallucinate $\ngen$ new examples using the hallucinator.
Each hallucinated example is obtained by sampling a real example $(x,y)$ from $\Strain$, sampling a noise vector $z$, and passing $x$ and $z$ to $G$ to obtain a generated example $(x', y)$ where $x' = G(x,z; \mathbf{w}_G)$.
We take the set of generated examples $\Strain^G$ and add it to the set of real examples to produce an augmented training set $\Strain^\mathrm{aug} = \Strain \cup \Strain^G$. 
We can now simply use this augmented training set to produce conditional probability estimates using $h$.
Note that the hallucinator parameters are kept fixed here; any learning that happens, happens within the classification algorithm $h$.

\paragraph{Meta-training the hallucinator:}
The goal of the hallucinator is to produce examples that help the classification algorithm learn a better classifier.
This goal differs from realism: realistic examples might still fail to capture the many modes of variation of visual concepts, while unrealistic hallucinations can still lead to a good decision boundary~\cite{dai2017good}. 
We therefore propose to directly train the hallucinator to support the classification algorithm by using meta-learning. 

As before, in each meta-training iteration, we sample $m$ classes from the set of all classes, and \emph{at most} $n$ examples per class.
Then, for each class, we use $G$ to generate $\ngen$ additional examples till there are exactly $\naug$ examples per class. 
Again, each hallucinated example is of the form $(x',y)$, where $x' = G(x, z; \mathbf{w}_G)$, $(x,y)$ is a sampled example from $\Strain$ and $z$ is a sampled noise vector.
These additional examples are added to the training set $\Strain$ to produce an augmented training set $\Strain^\mathrm{aug}$.
Then this augmented training set is fed to the classification algorithm $h$, to produce the final loss $\sum_{(x,y) \in \Stest} L(h(x, \Strain^\mathrm{aug}), y)$, where $\Strain^\mathrm{aug} = \Strain \cup \Strain^G$ and
$\Strain^G = \{(G(x_i, z_i; \mathbf{w}_G),y_i)_{i=1}^{\ngen} : (x_i, y_i) \in \Strain\}$.

To train the hallucinator $G$, we require that the classification algorithm $h(x, \Strain^\mathrm{aug}; \mathbf{w})$ is differentiable with respect to the elements in $\Strain^\mathrm{aug}$.
This is true for many meta-learning algorithms.
For example, in prototypical networks, $h$ will pass every example in the training set through a feature extractor, compute the class means in this feature space, and use the distances between the test point and the class means to estimate class probabilities.
If the feature extractor is differentiable, then the classification algorithm itself is differentiable with respect to the examples in the training set.
This allows us to back-propagate the final loss and update not just the parameters of the classification algorithm $h$, but also the parameters $\mathbf{w}_G$ of the hallucinator.
Figure~\ref{fig:2} shows a schematic of the entire process.

Using meta-learning to train the hallucinator and the classification algorithm has two benefits.
First, the hallucinator is directly trained to produce the kinds of hallucinations that are useful for class distinctions, removing the need to precisely tune realism or diversity, or the right modes of variation to hallucinate.
Second, the classification algorithm is trained jointly with the hallucinator, which enables it to make allowances for any errors in the hallucination.
Conversely, the hallucinator can spend its capacity on suppressing precisely those errors which throw the classification algorithm off.

Note that the training process is completely agnostic to the specific meta-learning algorithm used.
We will show in our experiments that our hallucinator provides significant gains irrespective of the meta-learner.

\section{Experimental Protocol}
\vspace{-0.1cm}
We use the benchmark proposed by Hariharan and Girshick~\cite{HariharanICCV2017}.
This benchmark captures more realistic scenarios than others based on handwritten characters~\cite{LakeScience2015} or low-resolution images~\cite{VinyalsNIPS2016}.
The benchmark is based on ImageNet images and subsets of ImageNet classes.
First, in the \emph{representation learning} phase, a convolutional neural network (ConvNet) based feature extractor is trained on one set of classes with thousands of examples per class; this set is called the ``base'' classes $\Cbase$.
Then, in the \emph{low-shot learning} phase, the recognition system encounters an additional set of ``novel'' classes $\Cnovel$ with  a small number of examples $n$ per class.
It also has access to the base class training set. 
The system has to now learn to recognize both the base and the novel classes.
It is tested on a test set containing examples from both sets of classes, and it needs to output labels in the joint label space $\Cbase \cup \Cnovel$. 
Hariharan and Girshick report the top-5 accuracy averaged over all classes, and also the top-5 accuracy averaged over just base-class examples, and the top-5 accuracy averaged over just novel-class examples.

\paragraph{Tradeoffs between base and novel classes:}
We observed that in this kind of \emph{joint} evaluation, different methods had very different performance tradeoffs between the novel and base class examples and yet achieved similar performance on average.
This makes it hard to meaningfully compare the performance of different methods on just the novel or just the base classes. Further, we found that by changing hyperparameter values of some meta-learners it was possible to achieve substantially different tradeoff points without substantively changing average performance. This means that hyperparameters can be tweaked to make novel class performance look better at the expense of base class performance (or vice versa).

One way to concretize this tradeoff is by incorporating a prior over base and novel classes.
Consider a classifier that gives a score $s_k(x)$ for every class $k$ given an image $x$.
Typically, one would convert these into probabilities by applying a softmax function:
\begin{equation}
\vspace{-0.1cm}
p_k(x) = p(y=k | x) = \frac{e^{s_k}}{\sum_j e^{s_j}}.
\vspace{-0.1cm}
\label{eq:unbiased}
\end{equation}
However, we may have some prior knowledge about the probability that an image belongs to the base classes $\Cbase$ or the novel classes $\Cnovel$.
Suppose that the prior probability that an image belongs to one of the novel classes is $\mu$.
Then, we can update Equation~\eqref{eq:unbiased} as follows:
\begin{align}
\vspace{-0.1cm}
p_k(x) &= p(y=k| x) \\
&= p(y=k | y \in \Cbase, x) p(y \in \Cbase| x) \nonumber\\
&\;\;\;\;+ p(y=k | y \in \Cnovel, x) p(y \in \Cnovel | x) \\
&= \frac{e^{s_k} \mathbf{I}[k \in \Cbase]}{\sum_j e^{s_j} \mathbf{I}[j \in \Cbase]} (1 - \mu) \nonumber\\
&\;\;\;\;+ \frac{e^{s_k} \mathbf{I}[k \in \Cnovel]}{\sum_j e^{s_j} \mathbf{I}[j \in \Cnovel]} \mu.
\vspace{-0.1cm}
\end{align}
The prior probability $\mu$ might be known beforehand, but can also be cross-validated to correct for inherent biases in the scores $s_k$. 
However, note that in some practical settings, one may not have a held-out set of categories to cross-validate.
Thus resilience to this prior is important.

\begin{figure}
\centering
\includegraphics[width=0.71\linewidth]{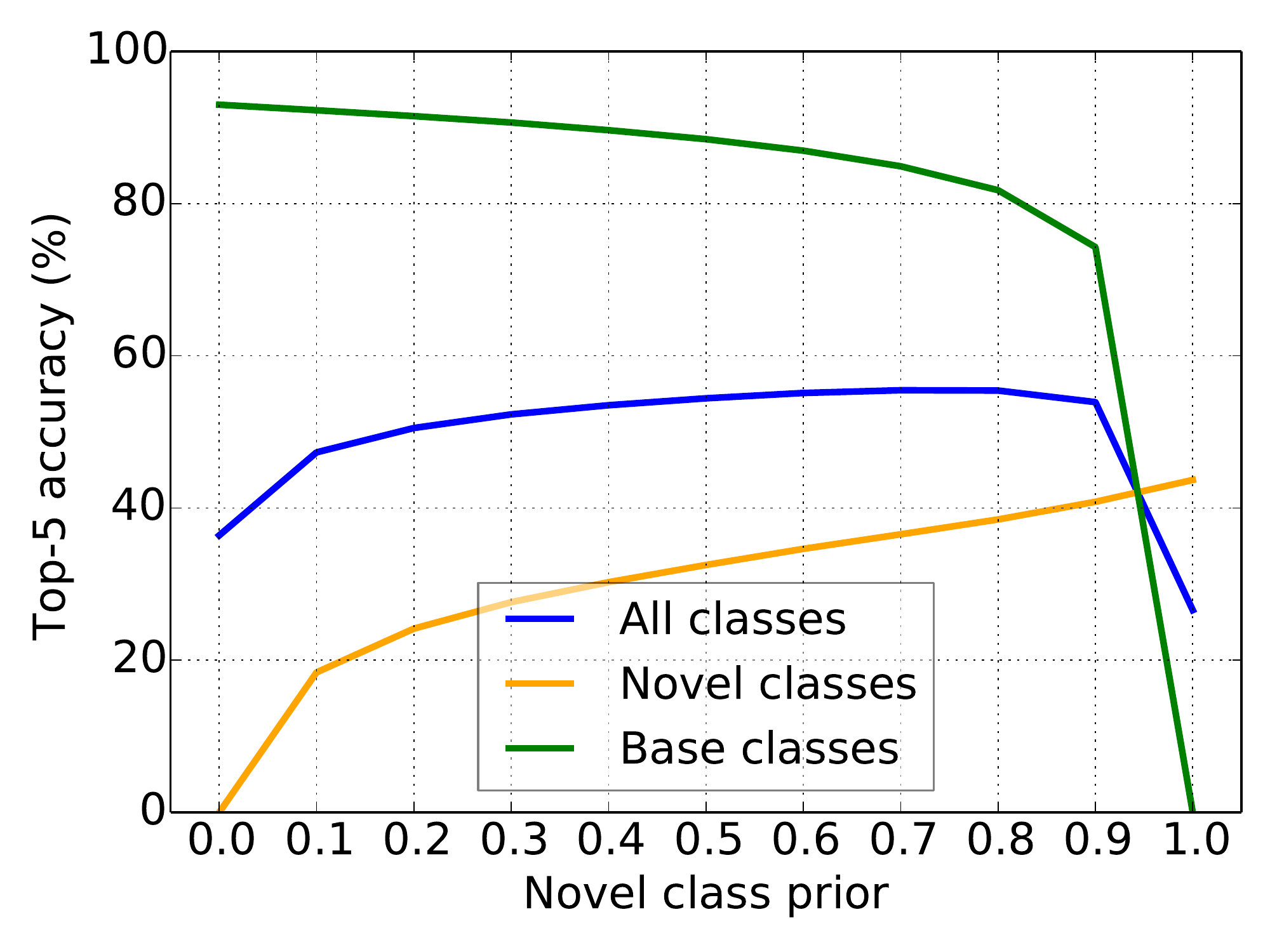}
\vspace{-0.2cm}
\caption{The variation of the overall, novel class, and base class accuracy for MN in the evaluation proposed by Hariharan and Girshick~\cite{HariharanICCV2017} as the novel class prior $\mu$ is varied. }
\label{fig:prior}
\vspace{-0.2cm}
\end{figure}

Figure~\ref{fig:prior} shows the impact of this prior on matching networks in the evaluation proposed by Hariharan and Girshick~\cite{HariharanICCV2017}.
Note that the overall accuracy remains fairly stable, even as novel class accuracy rises and base class accuracy falls.
Such prior probabilities for calibration were proposed for the zero-shot learning setting by Chao \etal~\cite{ChaoECCV2016}.

\paragraph{A new evaluation:}
The existence of this tunable tradeoff between base and novel classes makes it hard to make apples-to-apples comparisons of novel class performance if the model is tasked with making predictions in the joint label space.
Instead, we use a new evaluation protocol that evaluates four sets of numbers:
\begin{packed_enum}
\item The model is given test examples from the novel classes, and is only supposed to pick a label from the novel classes. That is, the label space is restricted to $\Cnovel$ (note that doing so is equivalent to setting $\mu=1$ for prototypical networks but not for matching networks and prototype matching networks because of the contextual embeddings). We report the top-5 accuracy on the novel classes in this setting.
\item Next, the model is given test examples from the base classes, and the label space is restricted to the base classes. We report the top-5 accuracy in this setting.
\item The model is given test examples from both the base and novel classes in equal proportion, and the model has to predict labels from the joint label space. We report the top-5 accuracy averaged across all examples. We present numbers both with and without a novel class prior $\mu$; the former set cross-validates $\mu$ to achieve the highest average top-5 accuracy.
\end{packed_enum}
Note that, following~\cite{HariharanICCV2017}, we use a disjoint set of classes for cross-validation and testing. This prevents hyperparameter choices for the hallucinator, meta-learner, and novel class prior from becoming overfit to the novel classes that are seen for the first time at test time.

\section{Experiments}
\subsection{Implementation Details}
\vspace{-0.1cm}
Unlike prior work on meta-learning which experiments with small images and few classes~\cite{VinyalsNIPS2016, SnellArxiv2017, FinnICML2017, RaviICLR2017}, we use high resolution images and our benchmark involves hundreds of classes.
This leads to some implementation challenges.
Each iteration of meta-learning at the very least has to compute features for the training set $\Strain$ and the test set $\Stest$.
If there are 100 classes with 10 examples each, then this amounts to 1000 images, which no longer fits in memory.
Training a modern deep convolutional network with tens of layers from scratch on a meta-learning objective may also lead to a hard learning problem.

Instead, we first train a convolutional network based feature extractor on a simple classification objective on the base classes $\Cbase$.
Then we extract and save these features to disk, and use these pre-computed features as inputs. For most experiments, consistent with~\cite{HariharanICCV2017}, we use a small ResNet-10 architecture~\cite{HeCVPR2016}.
Later, we show some experiments using the deeper ResNet-50 architecture~\cite{HeCVPR2016}.

\paragraph{Meta-learner architectures:} We focus on state-of-the-art meta-learning approaches, including prototypical networks (PN)~\cite{SnellArxiv2017}, matching networks (MN)~\cite{VinyalsNIPS2016}, and our improvement over MN --- prototype matching networks (PMN). For PN, the embedding architecture consists of two MLP layers with ReLU as the activation function. We use Euclidean distance as in~\cite{SnellArxiv2017}. For MN,  following~\cite{VinyalsNIPS2016}, the embedding architecture consists of a one layer bi-directional LSTM that embeds training examples and attention LSTM that embeds test samples. We use cosine distance as in~\cite{VinyalsNIPS2016}. For our PMN, we collapse every class to its class mean before the contextual embeddings of the test examples, and we keep other design choices the same as those in MN.

\paragraph{Hallucinator architecture and initialization:} For our hallucinator $G$, we use a three layer MLP with ReLU as the activation function. We add a ReLU at the end since the pre-trained features are known to be non-negative. 
All hidden layers have a dimensionality of 512 for ResNet-10 features and 2048 for ResNet-50 features. 
Inspired by~\cite{le2015simple}, we initialize the weights of our hallucinator network as block diagonal identity matrices. 
This significantly outperformed standard initialization methods like random Gaussian, since the hallucinator  can ``copy'' its seed examples to produce a reasonable generation immediately from initialization.

\subsection{Results}
\begin{table*}[t!]
\centering
\renewcommand{\arraystretch}{1.2}
\renewcommand{\tabcolsep}{1.2mm}
\resizebox{0.92\linewidth}{!}{\begin{tabular}{lccccc@{\hspace{8mm}}ccccc@{\hspace{8mm}}ccccc@{\hspace{8mm}}}
& \multicolumn{5}{c@{\hspace{8mm}}}{Novel} & \multicolumn{5}{c@{\hspace{8mm}}}{All} & \multicolumn{5}{c@{\hspace{8mm}}}{All with prior}\\
Method &$n$=1 & 2 & 5 & 10 & 20&$n$=1 & 2 & 5 & 10 & 20&$n$=1 & 2 & 5 & 10 & 20\\
\midrule
\emph{ResNet}-10 &  &  &  &  &  &  &  &  &  &  &  &  &  & \\
\;PMN w/ G* & \firstrank{45.8} & \firstrank{57.8} & \firstrank{69.0} & \firstrank{74.3} & \firstrank{77.4} & \firstrank{57.6} & \firstrank{64.7} & \firstrank{71.9} & \firstrank{75.2} & \firstrank{77.5} & \firstrank{56.4} & \firstrank{63.3} & \firstrank{70.6} & \firstrank{74.0} & 76.2\\
\;PMN* & 43.3 & 55.7 & \secondrank{68.4} & \secondrank{74.0} & \secondrank{77.0} & 55.8 & 63.1 & \secondrank{71.1} & \secondrank{75.0} & \secondrank{77.1} & 54.7 & 62.0 & \secondrank{70.2} & \secondrank{73.9} & 75.9\\
\;PN w/ G* & \secondrank{45.0} & \secondrank{55.9} & 67.3 & 73.0 & 76.5 & \secondrank{56.9} & \secondrank{63.2} & 70.6 & 74.5 & 76.5 & \secondrank{55.6} & \secondrank{62.1} & 69.3 & 73.1 & 75.4\\
\;PN~\cite{SnellArxiv2017} & 39.3 & 54.4 & 66.3 & 71.2 & 73.9 & 49.5 & 61.0 & 69.7 & 72.9 & 74.6 & 53.6 & 61.4 & 68.8 & 72.0 & 73.8\\
\;MN~\cite{VinyalsNIPS2016} & 43.6 & 54.0 & 66.0 & 72.5 & 76.9 & 54.4 & 61.0 & 69.0 & 73.7 & 76.5 & 54.5 & 60.7 & 68.2 & 72.6 & 75.6\\
\midrule
\;LogReg & 38.4 & 51.1 & 64.8 & 71.6 & 76.6 & 40.8 & 49.9 & 64.2 & 71.9 & 76.9 & 52.9 & 60.4 & 68.6 & 72.9 & \secondrank{76.3}\\
\;LogReg w/ Analogies~\cite{HariharanICCV2017} & 40.7 & 50.8 & 62.0 & 69.3 & 76.5 & 52.2 & 59.4 & 67.6 & 72.8 & 76.9 & 53.2 & 59.1 & 66.8 & 71.7 & \firstrank{76.3}\\
\midrule
\midrule
\emph{ResNet}-50 &  &  &  &  &  &  &  &  &  &  &  &  &  & \\
\;PMN w/ G* & \firstrank{54.7} & \firstrank{66.8} & \firstrank{77.4} & \firstrank{81.4} & \firstrank{83.8} & \firstrank{65.7} & \firstrank{73.5} & \firstrank{80.2} & \firstrank{82.8} & \firstrank{84.5} & \firstrank{64.4} & \firstrank{71.8} & \firstrank{78.7} & \firstrank{81.5} & \firstrank{83.3}\\
\;PMN* & 53.3 & \secondrank{65.2} & \secondrank{75.9} & 80.1 & 82.6 & 64.8 & \secondrank{72.1} & 78.8 & 81.7 & \secondrank{83.3} & 63.4 & \secondrank{70.8} & \secondrank{77.9} & \secondrank{80.9} & \secondrank{82.7}\\
\;PN w/ G* & \secondrank{53.9} & 65.2 & 75.7 & \secondrank{80.2} & \secondrank{82.8} & \secondrank{65.2} & 72.0 & \secondrank{78.9} & \secondrank{81.7} & 83.1 & \secondrank{63.9} & 70.5 & 77.5 & 80.6 & 82.4\\
\;PN~\cite{SnellArxiv2017} & 49.6 & 64.0 & 74.4 & 78.1 & 80.0 & 61.4 & 71.4 & 78.0 & 80.0 & 81.1 & 62.9 & 70.5 & 77.1 & 79.5 & 80.8\\
\;MN~\cite{VinyalsNIPS2016} & 53.5 & 63.5 & 72.7 & 77.4 & 81.2 & 64.9 & 71.0 & 77.0 & 80.2 & 82.7 & 63.8 & 69.9 & 75.9 & 79.3 & 81.9\\
\end{tabular}
}
\caption{Top-5 accuracy on the novel classes and on all classes (with and without priors) for different values of $n$. $^*$Our methods. PN:~Prototypical networks, MN:~Matching networks, PMN:~Prototype matching networks, LogReg:~Logistic regression. Methods with ``w/ G''  use a meta-learned hallucinator.}
\label{tab:everything}
\vspace{-0.2cm}
\end{table*}

As in~\cite{HariharanICCV2017}, we run five trials for each setting of $n$ (the number of examples per novel class) and present the average performance. Different approaches are comparably good for base classes, achieving 92\% top-5 accuracy. We focus more on novel classes since they are more important in low-shot learning.
Table~\ref{tab:everything} contains a summary of the top-5 accuracy for novel classes and for the joint space both with and without a cross-validated prior.
Standard deviations for all numbers are of the order of 0.2\%.
We discuss specific results, baselines, and ablations below.

\begin{figure}
\includegraphics[width=0.495\linewidth]{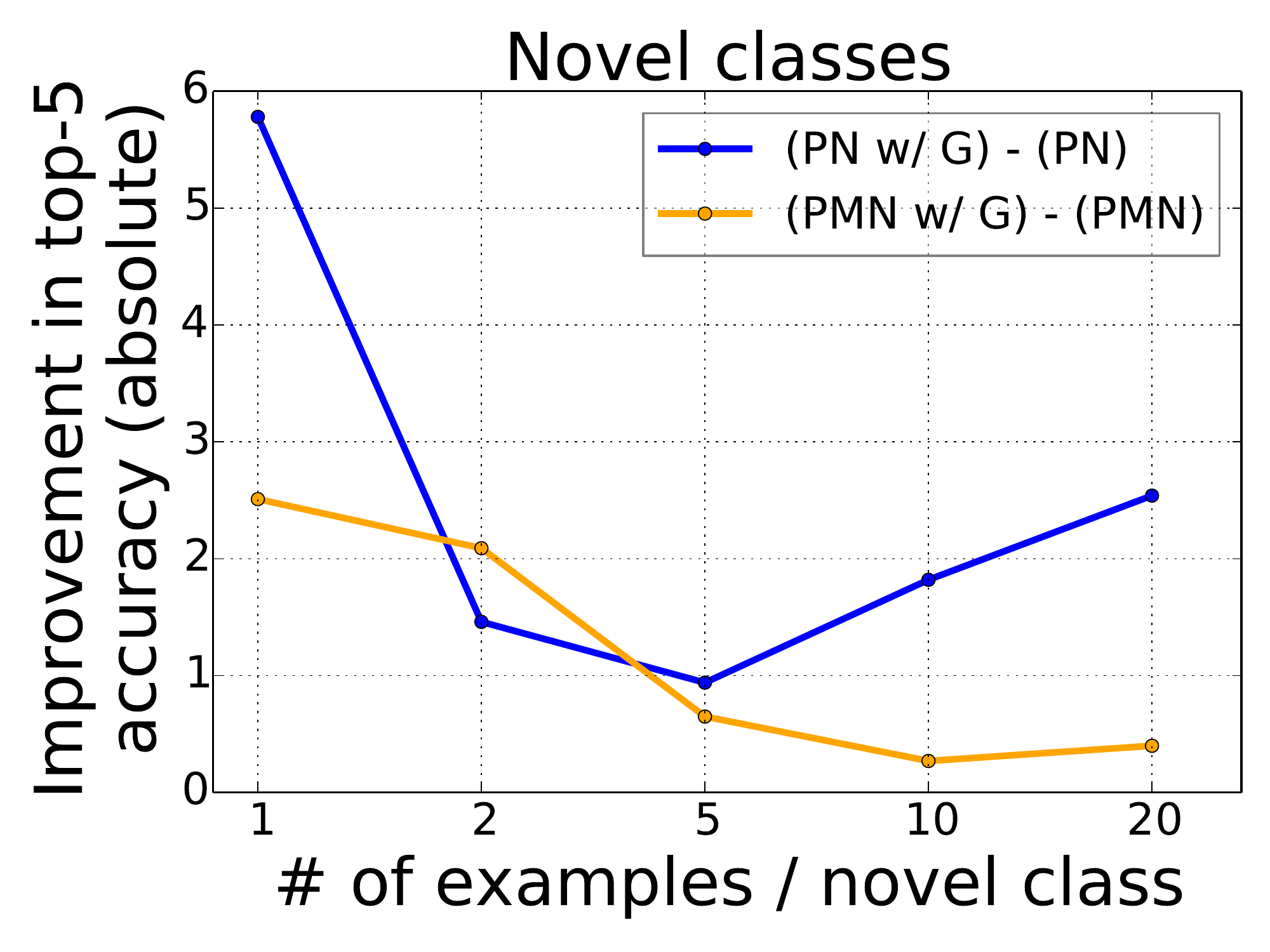}
\includegraphics[width=0.495\linewidth]{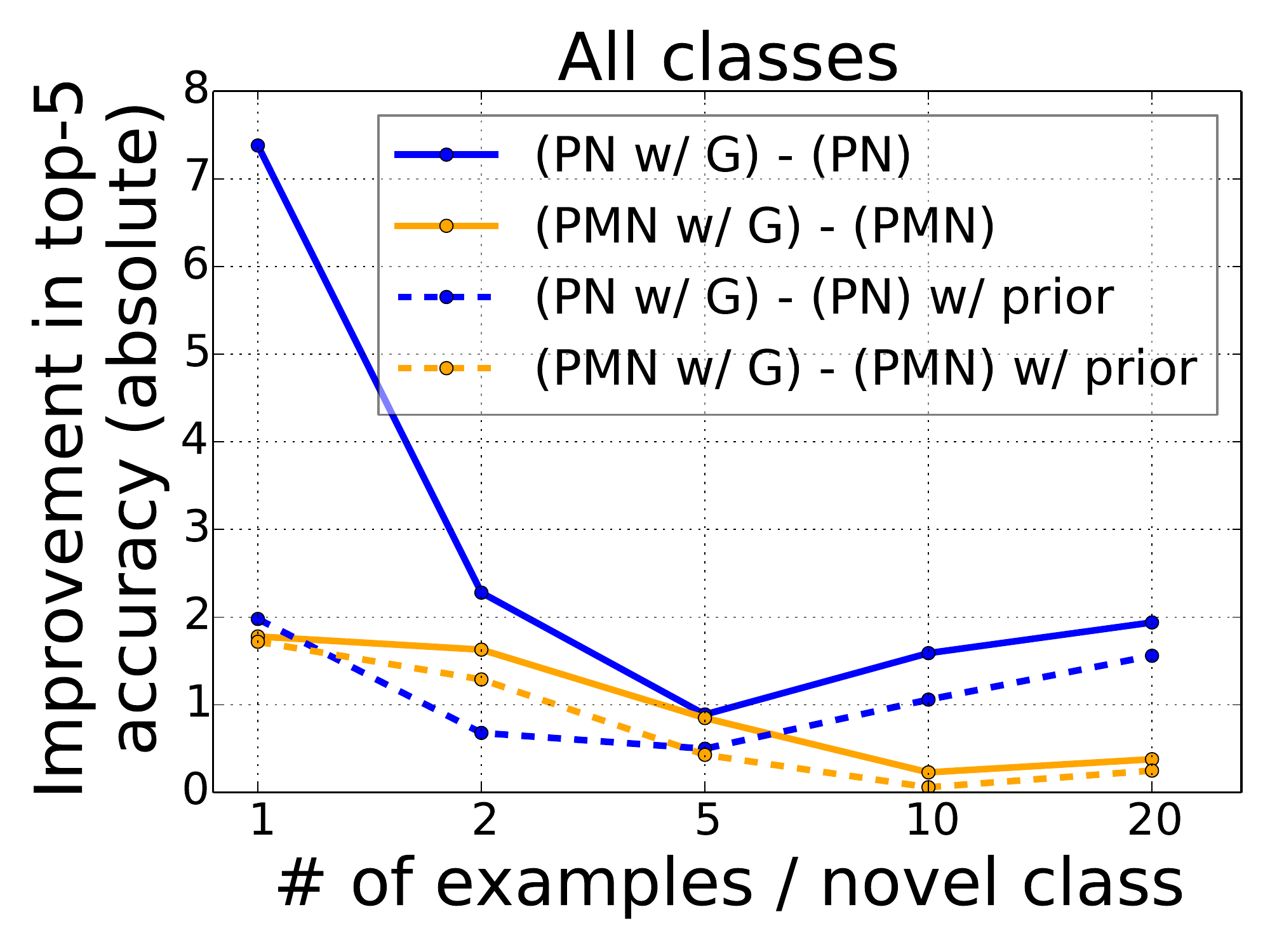}
\caption{Improvement in accuracy by learned hallucination for different meta-learners as a function of the number of examples available per novel class.}
\label{fig:hal_impact}
\vspace{-0.2cm}
\end{figure}

\begin{figure*}
\centering
\includegraphics[width=0.3\linewidth]{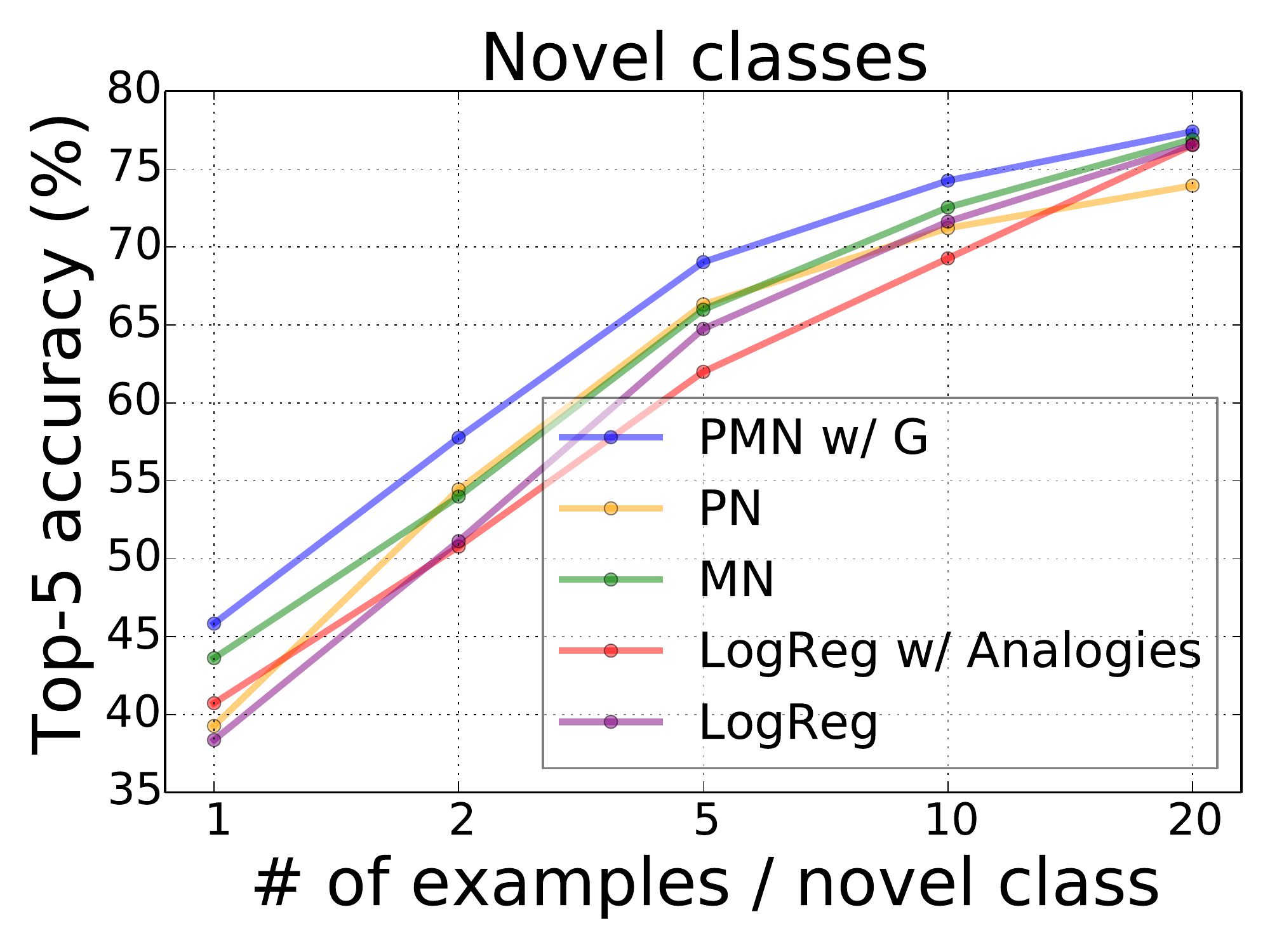}
\includegraphics[width=0.3\linewidth]{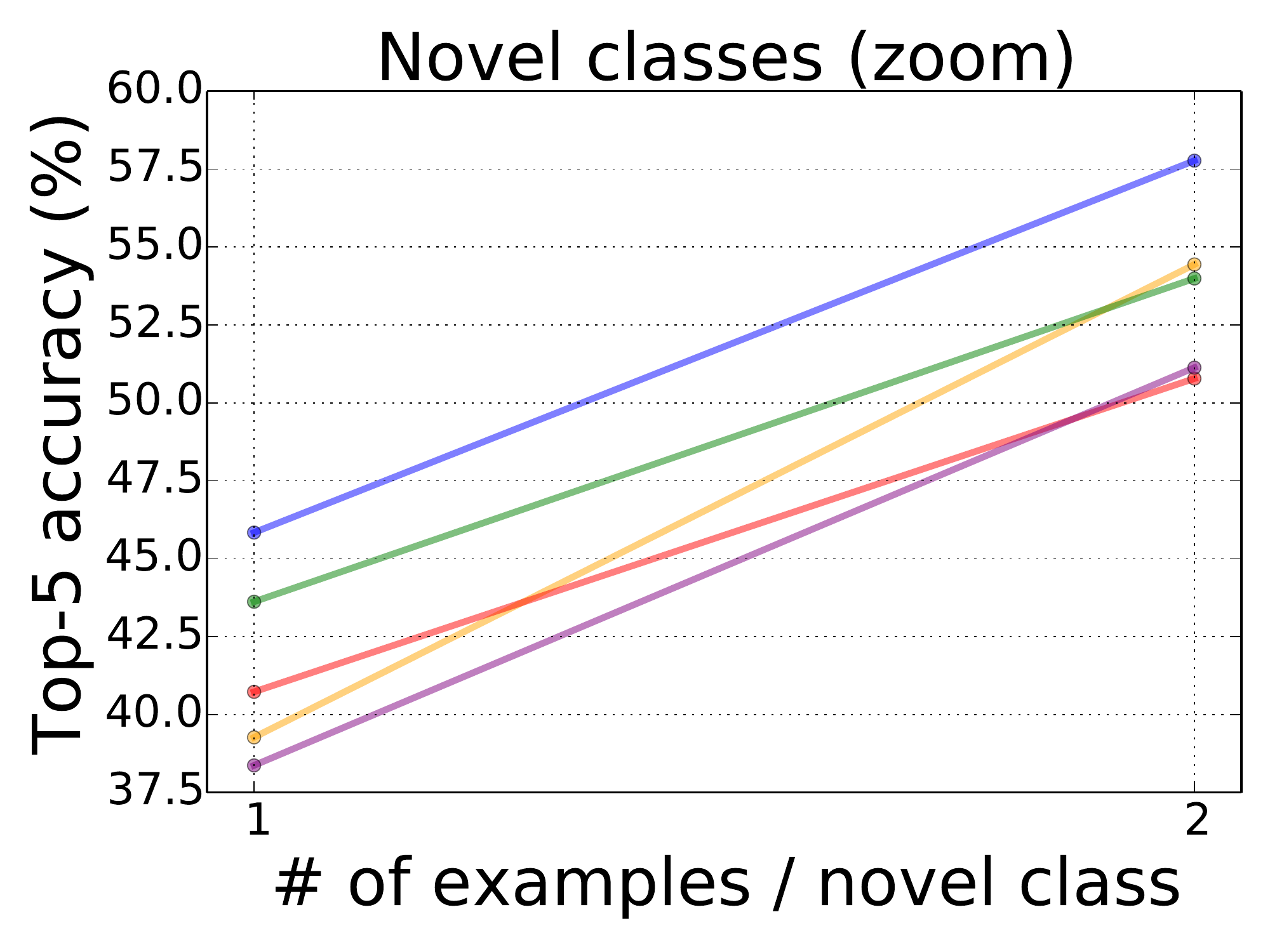}
\includegraphics[width=0.3\linewidth]{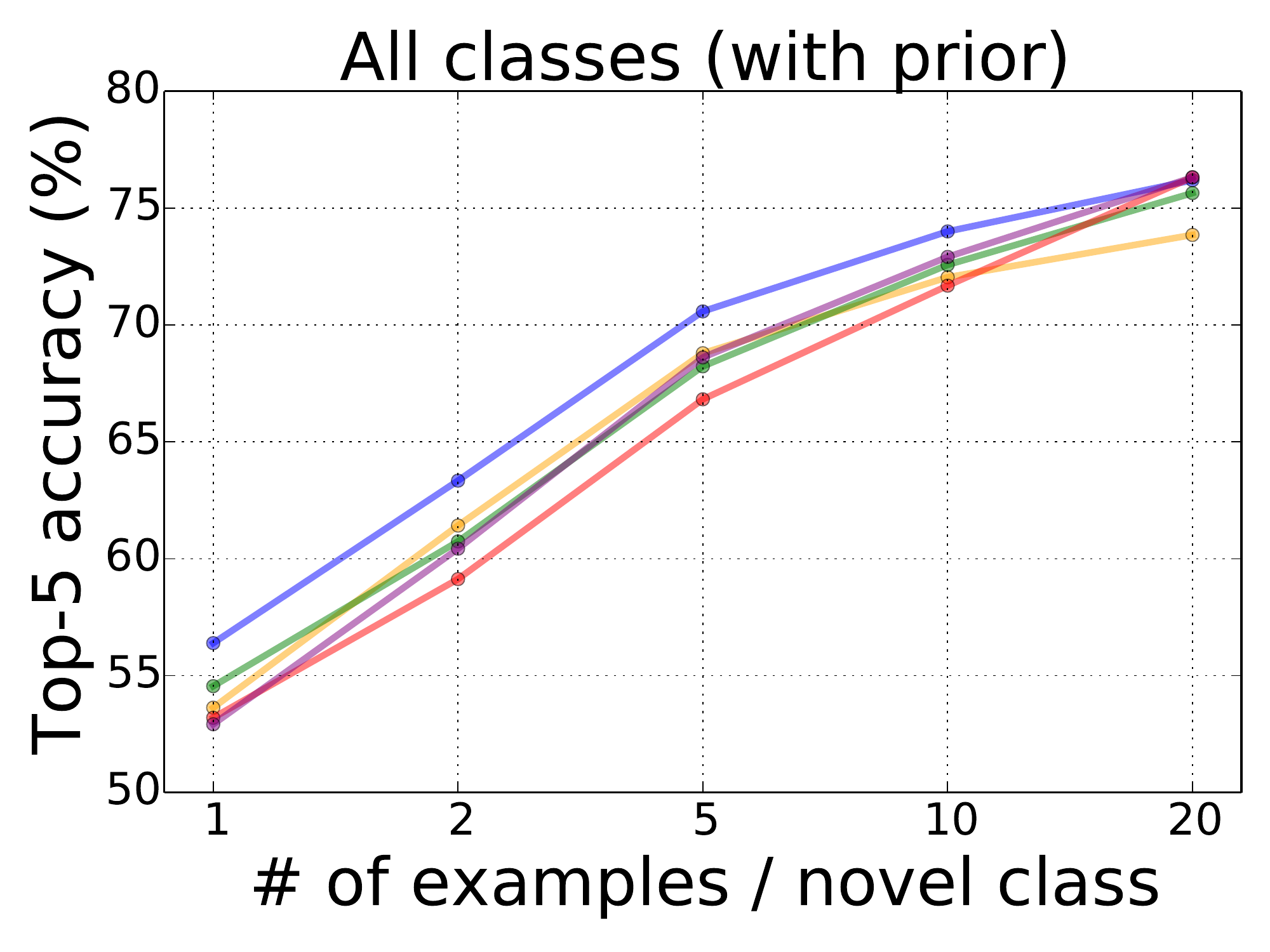}
\caption{Our best approach compared to previously published methods. From left to right: just the novel classes, zoomed in performance for the case when the number of examples per novel class $n \le 2$, performance on the joint label space with a cross-validated prior.}
\label{fig:baselines}
\vspace{-0.2cm}
\end{figure*}

\paragraph{Impact of hallucination:} We first compare meta-learners with and without hallucination to judge the impact of hallucination.
We look at prototypical networks (PN) and prototype matching networks (PMN) for this comparison.
Figure~\ref{fig:hal_impact} shows the improvement in top-5 accuracy we get from hallucination on top of the original meta-learner performance.
The  actual numbers are shown in Table~\ref{tab:everything}.

We find that our hallucination strategy improves novel class accuracy significantly, by up to 6 points for prototypical networks and 2 points for prototype matching networks.
This suggests that our approach is general and can work with different meta-learners.
While the improvement drops when more novel category training examples become available, the gains remain significant until $n=20$ for prototypical networks and $n=5$ for prototype matching networks.

Accuracy in the joint label space (right half of Figure~\ref{fig:hal_impact}) shows the same trend.
However, note that the gains from hallucination decrease significantly when we cross-validate for an appropriate novel-class prior $\mu$ (shown in dotted lines).
This suggests that part of the effect of hallucination is to provide resilience to mis-calibration.
This is important in practice where it might not be possible to do extensive cross-validation; in this case, meta-learners with hallucination demonstrate significantly higher accuracy than their counterparts without hallucination.

\paragraph{Comparison to prior work:} Figure~\ref{fig:baselines} and Table~\ref{tab:everything} compare our best approach (prototype matching networks with hallucination) with previously published approaches in low-shot learning.
These include prototypical networks~\cite{SnellArxiv2017}, matching networks~\cite{VinyalsNIPS2016}, and the following baselines:
\begin{packed_enum}
\item \emph{Logistic regression}: This baseline simply trains a linear classifier on top of a pre-trained ConvNet-based feature extractor that was trained on the base classes.
\item \emph{Logistic regression with analogies}: This baseline uses the procedure described by Hariharan and Girshick~\cite{HariharanICCV2017} to hallucinate additional examples. 
These additional examples are added to the training set and used to train the linear classifier.
\end{packed_enum} 

Our approach easily outperforms all baselines, providing almost a 2 point improvement across the board on the novel classes, and similar improvements in the joint label space even after allowing for cross-validation of the novel category prior.
Our approach is thus state-of-the-art.

Another intriguing finding is that our proposed prototype matching network outperforms matching networks on novel classes as more novel class examples become available (Table~\ref{tab:everything}).
On the joint label space, prototype matching networks are better across the board.

Interestingly, the method proposed by Hariharan and Girshick~\cite{HariharanICCV2017} \emph{underperforms} the standard logistic regression baseline (although it does show gains when the novel class prior is not cross-validated, as shown in Table~\ref{tab:everything}, indicating that its main impact is resilience to mis-calibration).

\paragraph{Unpacking the performance gain:}
To unpack where our performance gain is coming from, we perform a series of ablations to answer the following questions.

\begin{figure}[t]
\includegraphics[width=0.495\linewidth]{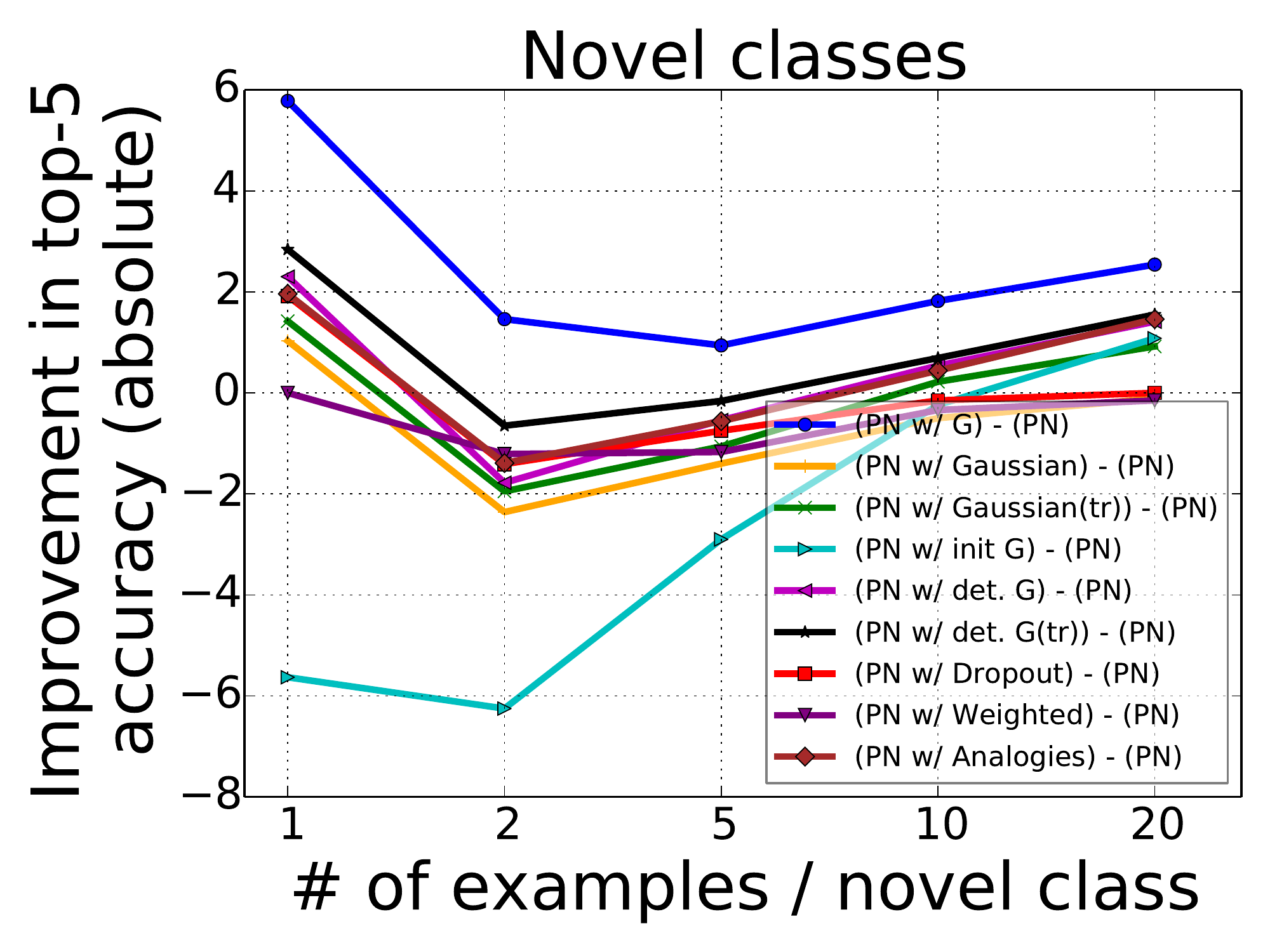}
\includegraphics[width=0.495\linewidth]{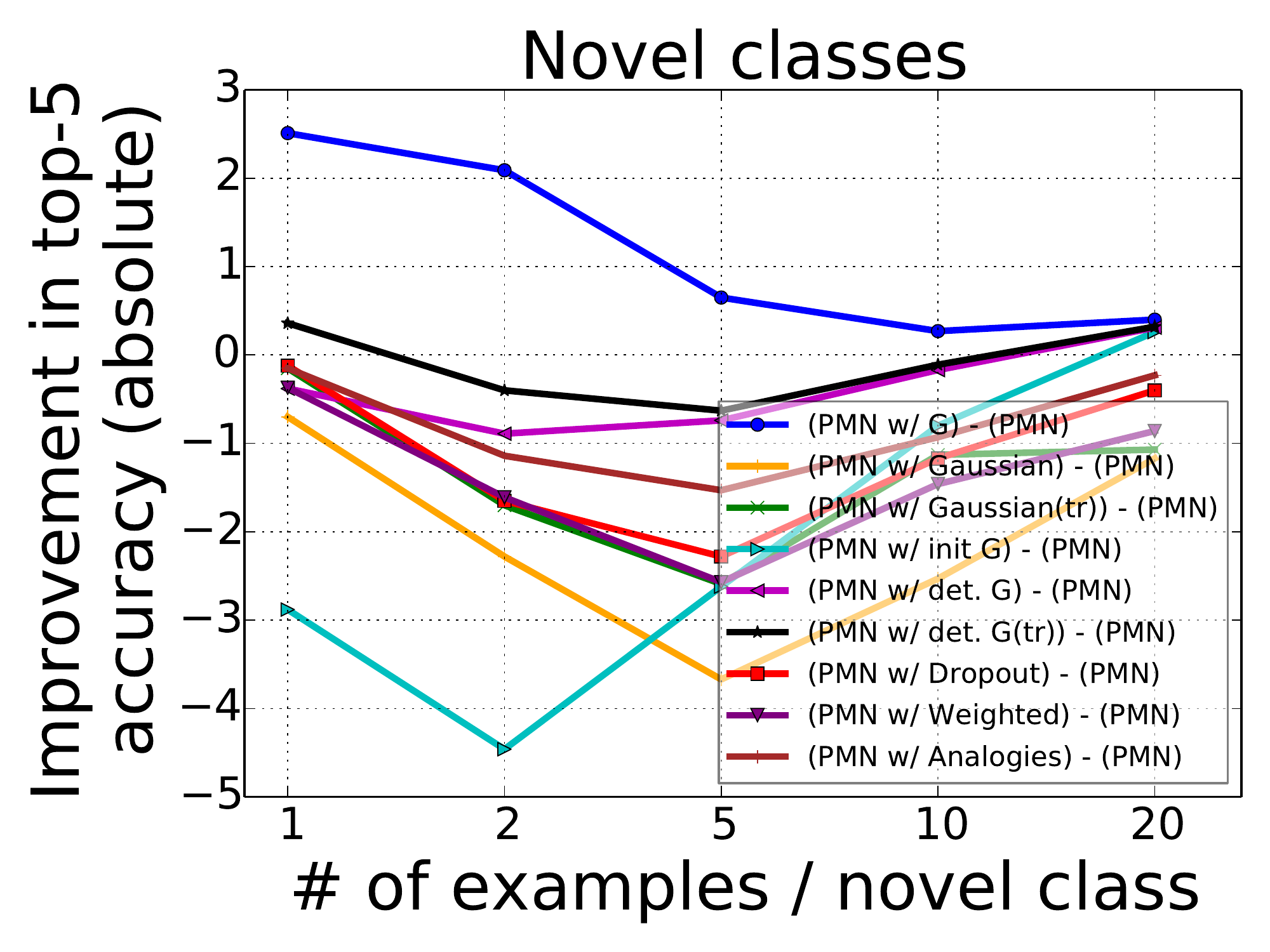}
\caption{Comparison of our learned hallucination with several ablations for both PN (left) and PMN (right). Our approach significantly outperforms the baselines, showing that a meta-learned hallucinator is important. {\bf Best viewed in color with zoom.}
}
\label{fig:ablation}
\vspace{-.2cm}
\end{figure}

\noindent\emph{Are sophisticated hallucination architectures necessary?}\\
In the semantic feature space learned by a convolutional network, a simple jittering of the training examples might be enough.
We created several baseline hallucinators that did such jittering by: (a) adding Gaussian noise with a diagonal covariance matrix estimated from feature vectors from the base classes, (b) using dropout (PN/PMN w/ Dropout), and (c) generating new examples through a weighted average of real ones (PN/PMN w/ Weighted). For the Gaussian hallucinator, we evaluated both a covariance matrix shared across classes and class-specific covariances. We found that the shared covariance outperformed class-specific covariances by 0.7 point and reported the best results. We tried both retraining the meta-learner with this Gaussian hallucinator, and using a pre-trained meta-learner: PN/PMN w/ Gaussian uses a pre-trained meta-learner and PN/PMN w/ Gaussian(tr) retrains the meta-learner. As shown in Figure~\ref{fig:ablation}, while such hallucinations help a little, they often hurt significantly, and lag the accuracy of our approach by at least 3 points.
This shows that generating useful hallucinations is not easy and requires sophisticated architectures.

\noindent\emph{Is meta-learning the hallucinator necessary?} \\
Simply passing Gaussian noise through an untrained convolutional network can produce complex distributions.
In particular, ReLU activations might ensure the hallucinations are non-negative, like the real examples.
We compared hallucinations with (a) an \emph{untrained} $G$ and (b) a \emph{pre-trained and fixed} $G$ based on analogies from~\cite{HariharanICCV2017} with our meta-trained version to see the impact of our training.
Figure~\ref{fig:ablation} shows the impact of these baseline hallucinators (labeled PN/PMN w/ init G and PN/PMN w/ Analogies, respectively).
These baselines hurt accuracy significantly, suggesting that meta-training the hallucinator is important.

\noindent\emph{Does the hallucinator produce diverse outputs?} \\
A persistent problem with generative models is that they fail to capture multiple modes~\cite{SalimansNIPS2016}.
If this is the case, then any one hallucination should look very much like the others, and simply replicating a single hallucination should be enough.
We compared our approach with: (a) a deterministic baseline that uses our trained hallucinator, but simply uses a fixed noise vector $z = 0$ (PN/PMN w/ det. G) and (b) a baseline that  uses replicated hallucinations during both training and testing (PN/PMN w/ det. G(tr)).
These baselines had a very small, but negative effect.
This suggests that our hallucinator produces \emph{useful, diverse} samples.

\begin{figure}
\centering
\begin{subfigure}[b]{0.453\linewidth}
\includegraphics[width=\linewidth]{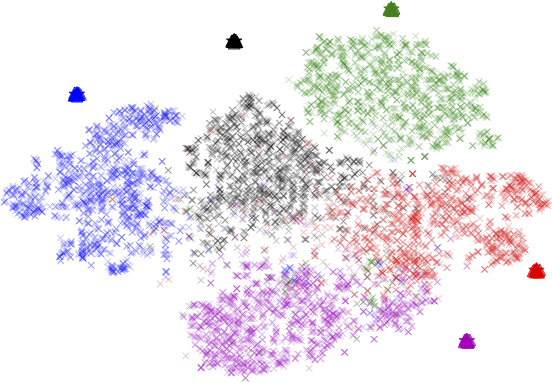}
\caption{Gaussian baseline}
\end{subfigure}
\begin{subfigure}[b]{0.453\linewidth}
\includegraphics[width=\linewidth]{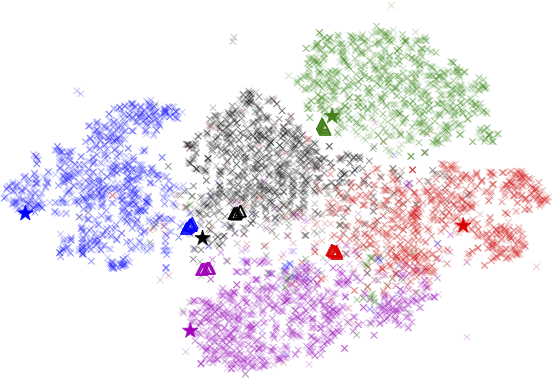}
\caption{$G$ with 1 seed}
\end{subfigure}\\
\begin{subfigure}[b]{0.453\linewidth}
\includegraphics[width=\linewidth]{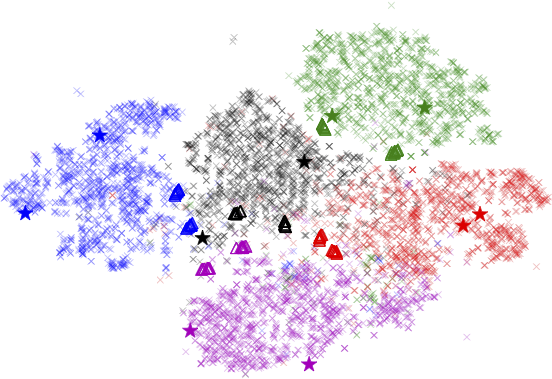}
\caption{2 seeds}
\end{subfigure}
\begin{subfigure}[b]{0.453\linewidth}
\includegraphics[width=\linewidth]{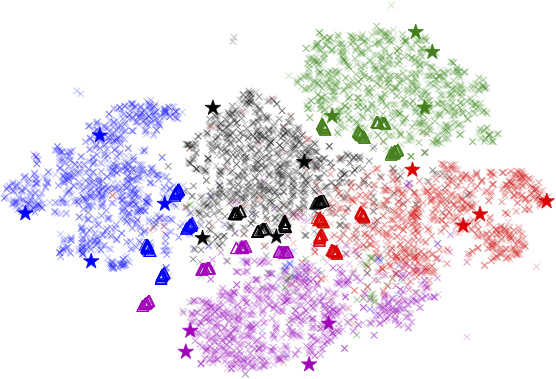}
\caption{4 seeds}
\end{subfigure}
\caption{t-SNE visualizations of hallucinated examples. Seeds are shown as stars, real examples as crosses, hallucinations as triangles. (a) Gaussian, single seed. (b,c,d) Our approach, 1, 2, and 4 seeds respectively. \textbf{Best viewed in color with zoom.}}
\label{fig:t-sne}
\vspace{-0.5cm}
\end{figure}

\paragraph{Visualizing the learned hallucinations:} Figure~\ref{fig:t-sne} shows t-SNE~\cite{maaten2008visualizing} visualizations of hallucinated examples for novel classes from our learned hallucinator and a baseline Gaussian hallucinator for prototypical networks. As before, we used statistics from the base class distribution for the Gaussian hallucinator.
Note that t-SNE tends to expand out parts of the space where examples are heavily clustered together.
Thus, the fact that the cloud of hallucinations for the Gaussian hallucinator is pulled away from the class distributions suggests that these hallucinations are very close to each other and far away from the rest of the class.
In contrast, our hallucinator matches the class distributions more closely, and with different seed examples captures different parts of the space.  
Interestingly, our generated examples tend to cluster around the class boundaries. 
This might be an artifact of t-SNE, or perhaps a consequence of discriminative training of the hallucinator.
However, our hallucinations are still fairly clustered; increasing the diversity of these hallucinations is an avenue for future work.

\paragraph{Representations from deeper models:}
All experiments till now used a feature representation trained using the ResNet-10 architecture~\cite{HariharanICCV2017}.
The bottom half of Table~\ref{tab:everything} shows the results on features from a ResNet-50 architecture.
As expected, all accuracies are higher, but our hallucination strategy still provides gains on top of both prototypical networks and prototype matching networks.

\section{Conclusion}
\vspace{-0.1cm}
In this paper, we have presented an approach to low-shot learning that uses a trained hallucinator to generate additional examples.
Our hallucinator is trained end-to-end with meta-learning, and we show significant gains on top of multiple meta-learning methods.
Our best proposed model achieves state-of-the-art performance on a realistic benchmark by a comfortable margin.
Future work involves pinning down exactly the effect of the hallucinated examples.

{\footnotesize {\bf Acknowledgments:} We thank Liangyan Gui, Larry Zitnick, Piotr Doll\'ar, Kaiming He, and Georgia Gkioxari for valuable and insightful discussions. This work was supported in part by ONR MURI N000141612007 and U.S. Army Research Laboratory (ARL) under the Collaborative Technology Alliance Program, Cooperative Agreement W911NF-10-2-0016. We also thank NVIDIA for donating GPUs and AWS Cloud Credits for Research program.}

{\small
\bibliographystyle{ieee}
\bibliography{main}
}

\end{document}